\pdfoutput=1

\documentclass[11pt]{article}

\usepackage[final]{coling}

\usepackage{times}
\usepackage{latexsym}

\usepackage[T1]{fontenc}

\usepackage[utf8]{inputenc}

\usepackage{microtype}

\usepackage{inconsolata}

\usepackage{graphicx}

\title{Towards Building Efficient Sentence BERT Models using Layer Pruning}

\author{Anushka Shelke$^{1,3}$, Riya Savant$^{1,3}$, Raviraj Joshi$^{2,3}$ \\
$^1$MKSSS Cummins College of Engineering for Women, Pune \\
$^2$Indian Institute of Technology Madras \\
$^3$L3Cube Labs, Pune \\
ravirajoshi@gmail.com}

\begin{document}
\maketitle
\begin{abstract}
This study examines the effectiveness of layer pruning in creating efficient Sentence BERT (SBERT) models. Our goal is to create smaller sentence embedding models that reduce complexity while maintaining strong embedding similarity. We assess BERT models like Muril and MahaBERT-v2 before and after pruning, comparing them with smaller, scratch-trained models like MahaBERT-Small and MahaBERT-Smaller.
Through a two-phase SBERT fine-tuning process involving Natural Language Inference (NLI) and Semantic Textual Similarity (STS), we evaluate the impact of layer reduction on embedding quality. Our findings show that pruned models, despite fewer layers, perform competitively with fully layered versions. Moreover, pruned models consistently outperform similarly sized, scratch-trained models, establishing layer pruning as an effective strategy for creating smaller, efficient embedding models.
These results highlight layer pruning as a practical approach for reducing computational demand while preserving high-quality embeddings, making SBERT models more accessible for languages with limited technological resources.
\end{abstract}

\section{Introduction}

Language models have evolved significantly in recent years. Although RNNs were once popular, they lack context embedding. Transformers \cite{vaswani2023attention} have emerged as superior, offering parallel processing for faster sequence handling and greater memory efficiency by utilizing position embeddings. Notably, BERT \cite{devlin2018bert}, a leading language model, adopts the Transformer architecture, significantly improving performance across a range of NLP tasks by capturing deep contextual relationships within text. 

BERT's architecture is built upon a multi-layer bidirectional Transformer encoder, drawing from the foundational framework of transformers \cite{vaswani2023attention}.$BERT_{BASE}$
\cite{devlin2018bert} is endowed with 110 million parameters, whereas $BERT_{LARGE}$
boasts 340 million parameters. The deployment of BERT models remains challenging in resource-constrained environments typical of many low-resource languages due to their substantial computational demands. 



While BERT excels at capturing contextualized word embeddings, it doesn't directly provide sentence-level representations. SBERT \cite{reimers2019sentencebert} addresses this limitation by modifying BERT's architecture to efficiently generate sentence embeddings. SBERT accomplishes this through the use of siamese and triplet network structures.The modification introduced by SBERT makes the BERT model more complex by extending its capabilities beyond word-level embeddings to include sentence-level representations. This added complexity enables BERT to capture higher-level semantic information and relationships between entire sentences, enhancing its utility in a wider range of natural language processing tasks.

These fine-tuned BERT models, with their large number of parameters, present challenges for low-capability devices or applications with strict latency requirements due to their resource-intensive nature. Various model compression techniques, including pruning, quantization, knowledge distillation, and architectural modifications, have been employed on BERT \cite{Ganesh_2021} to decrease the model size and computational demands, thereby increasing computation latency.


Building on the efforts to address the challenges posed by resource-intensive BERT models, our research delves into reducing the complexity of SBERT models without compromising performance. 
Layer pruning, which involves selectively removing less critical parts of the neural network, offers a promising solution for enhancing the efficiency of SBERT models. This is especially important for processing languages within environments constrained by limited computing infrastructure.

Model pruning, specifically layer pruning, seeks to address the inefficiencies related to the size and complexity of models like BERT, SBERT. The objective is to reduce the model's size and computational demands while maintaining or enhancing its performance. Techniques vary from removing individual neurons to whole layers. In the context of transformer-based models, a study \cite{fan2019reducing} demonstrated that strategic layer removal could reduce model size substantially with minimal impact on performance.




In our research, we delve into recent developments in adapting Sentence-BERT (SBERT) models for low-resource languages, focusing particularly on Marathi and Hindi. 
The L3Cube-MahaSBERT and HindSBERT \cite{joshi2022l3cubemahasbert} models were established as benchmarks for generating high-quality sentence embeddings in Marathi and Hindi, respectively.
These specialized models are highlighted for their effectiveness in processing these low-resource languages. These models have been rigorously trained and evaluated across various NLP tasks, including text classification and semantic similarity.

\begin{figure}[t]
    \centering
    \includegraphics[width=\linewidth]{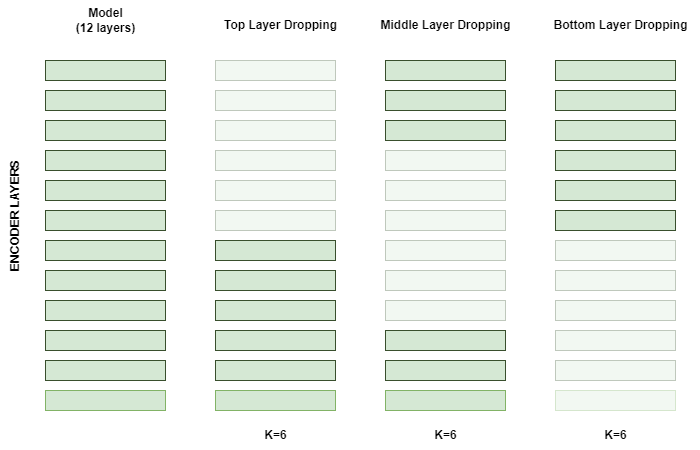} 
    \caption{Layer Pruning Strategies. Top-layer pruning performs the best for training SBERT models.}
    \label{fig:pruning strategies}
\end{figure}

Our research aims to extend these foundational models by applying layer-pruning techniques to enhance their efficiency without compromising the quality of the embeddings. By integrating layer pruning, we seek to reduce the computational demand and improve the operational feasibility of deploying SBERT models in real-world applications, making advanced NLP tools more accessible for languages that traditionally have fewer technological resources.
\begin{itemize}
\item A research  \cite{sajjad2022effect} has showcased a range of layer pruning strategies, underscoring their effectiveness. These techniques maintain an impressive 98\% of the original performance even after removing 40\% of the layers from BERT, RoBERTa, and XLNet models. 
\item Expanding upon these findings, we applied several layer pruning methods—such as top-layer pruning, middle-layer pruning, and bottom-layer pruning—to SBERT models, as illustrated in the accompanying figure \ref{fig:pruning strategies}. In this context, the parameter "k" represents the number of layers removed from the original model. 
\item After evaluating all three approaches, we discovered that top-layer pruning yielded the best performance. Therefore, we chose top-layer pruning for our subsequent experiments. To further test the performance of these pruned models, we fine-tuned them using NLI+STS training.
\item We compare 2-layer and 6-layer models created through layer pruning of MahaBERT-v2 with similar-sized models trained from scratch, such as MahaBERT-Small and MahaBERT-Smaller. Our observations show that the pruned models consistently outperform the scratch-trained models. Therefore, we recommend layer pruning followed by SBERT-like fine-tuning to create smaller embedding models, rather than training smaller models from scratch and then applying SBERT-like fine-tuning, which is highly computationally intensive.
\item Remarkably, these fine-tuned pruned models demonstrate competitive performance compared to larger models, despite being 50\% to 80\% smaller in size.
\end{itemize}

\section{Related Work}
This section discusses the progression of transformer-based models, with a specific focus on their optimization for enhanced efficiency and application in resource-constrained environments.

Introduced by \cite{devlin2019bert} BERT revolutionized NLP tasks by employing a bidirectional training of Transformer, a novel architecture that was originally used in the paper \cite{vaswani2023attention} thereby encapsulating a deeper contextual understanding. The paper \cite{reimers2019sentencebert} introduces Sentence-BERT (SBERT), a modification of the original BERT model that uses Siamese and triplet network structures to efficiently generate sentence embeddings for enhanced performance in semantic similarity tasks.

\cite{zhu2017prune} evaluates the impact of different pruning techniques on neural network compression and performance across various models and tasks. As discussed in their  \cite{fan2019reducing}, it has been shown that carefully targeted removal of layers can significantly decrease the size of a model while having only a minimal effect on its performance. Furthermore, the study by \cite{michel2019sixteen}, titled "Are Sixteen Heads Really Better than One?" shows that many attention heads in transformers can be pruned without significant degradation in capabilities, highlighting the redundancy in these models.

We explore research aimed at enhancing the efficiency of transformer models, particularly through model compression techniques. 
Key studies in this area include \cite{hubara2016quantized} and \cite{jiao2020tinybert}, which provide valuable insights into designing more efficient models without significant loss in performance.
The main goal of TinyBERT is to distill the knowledge from a large pre-trained language model, such as BERT, into a smaller model, while maintaining performance.

Additionally, we delve into the literature on layer pruning techniques, which specifically address methods for optimizing neural network architectures by identifying and removing redundant or less important layers. 
In this domain, valuable strategies have been employed for reducing the computational
burden of neural network models through systematic layer
pruning approaches \cite{liu2017learning}.
An iterative algorithm \cite{pietron2020retrain} is introduced for layer pruning, reducing storage demands in pre-trained neural networks. It selects layers based on complexity and sensitivity, applying reverse pruning if accuracy drops. 

Layer pruning reduces resource usage in CNNs by eliminating entire layers based on their importance estimated through PLS projection \cite{9007458}. 
It can be followed by filter-oriented pruning for additional compression. Structured pruning \cite{He_2024} encompasses a range of techniques such as filter ranking methods,  dynamic execution, the lottery ticket hypothesis, etc. Layer-wise pruning ratios extend traditional weight pruning strategies by focusing on determining the optimal pruning rate for each layer. 

Another method for layer-wise pruning based on feature representations \cite{8485719} is introduced. Unlike conventional methods that prune based on weight information, this approach identifies redundant parameters by examining the features learned in convolutional layers, operating at a layer level. A novel approach called layer-compensated pruning \cite{chin2018layercompensated} incorporates meta-learning to address both how many filters to prune per layer and which filters to prune. Tests on ResNet and MobileNetV2 networks across multiple datasets validate the algorithm's effectiveness.

\section{Methodologies}

\begin{figure*}[t]
    \centering
    \includegraphics[width=\linewidth]{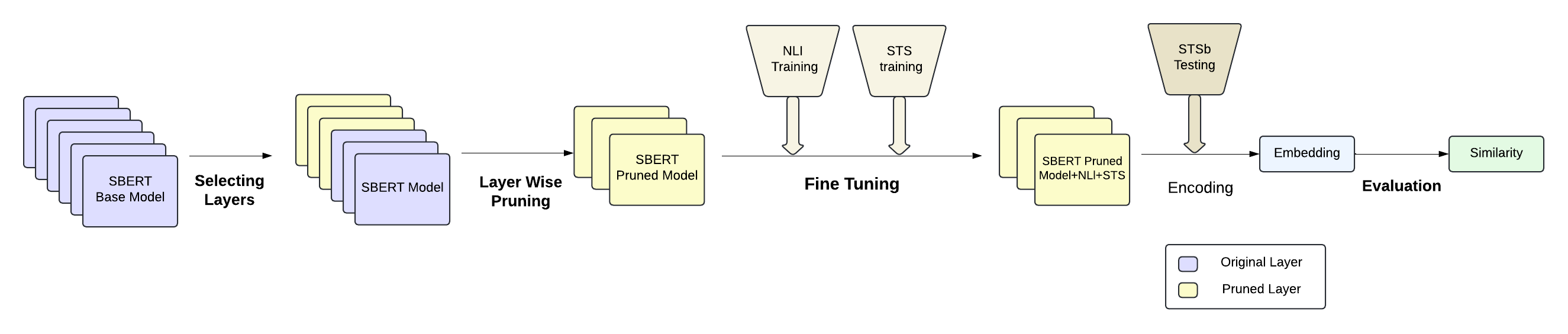}
    \caption{Layer Pruning on SBERT model}
    \label{fig:Experiment}
\end{figure*}

SBERT models are known for their complexity and large size. Fig. \ref{fig:Experiment} depicts the process of training a smaller SBERT (Sentence-BERT) model using a technique known as layer pruning. Starting with the original SBERT base model, which consists of multiple layers, the process involves systematically removing certain layers to create a pruned version of the model. This layer-wise pruning aims to reduce the model's complexity without significantly compromising its performance.

Our initial experiments focused on identifying the most effective layer-pruning strategy to optimize the model's performance. We explored several pruning methods, including top-layer pruning, middle-layer pruning, and bottom-layer pruning as shown in \ref{fig:pruning strategies}, to evaluate their impact on model's efficiency and accuracy. Each strategy was tested by removing a specified number of layers, denoted by the parameter "k", from different positions in the model. This approach allowed us to systematically assess how the removal of layers affected the overall performance and computational requirements.

The pruned model is then fine-tuned through two specialized training phases: Natural Language Inference (NLI) training and Semantic Textual Similarity (STS) training. NLI training improves the model's ability to understand logical relationships between sentence pairs, categorizing them as entailment, contradiction, or neutral, whereas STS training focuses on assigning similarity scores to sentence pairs, enhancing the model’s ability to gauge semantic closeness. By integrating NLI pre-training and STS fine-tuning, a robust training framework is established for SBERT models.

Following the fine-tuning, the pruned model which is integrated with NLI and STS training is tested for its performance on the Semantic Textual Similarity benchmarks (STSb testing) dataset. This phase evaluates how effectively the model calculates the similarity between sentences. The final steps involve encoding these sentences into embeddings and evaluating their similarity and assessing the pruned model's accuracy and efficiency. Thus Fig.\ref{fig:Experiment} depicts a clear pathway from model complexity reduction through pruning to performance evaluation via embedding and similarity assessments.

\subsection{Dataset} 
\setcounter{subsubsection}{0}
\subsubsection{IndicXNLI} 
IndicXNLI\footnote{\url{https://github.com/divyanshuaggarwal/IndicXNLI}} comprises data from the English XNLI dataset that has been translated into eleven Indic languages including Marathi.\cite{aggarwal2022indicxnli} This includes translation of the training (392,702 entries), validation (2,490 entries), and evaluation sets (5,010 entries) from English into each of the eleven languages. From the IndicXNLI dataset, the training samples specific to each language are used to train the MahaSBERT models.

\subsubsection{STS benchmark(STSb)}
It comprises data from the English XNLI dataset that has been translated into eleven Indic languages including Marathi\footnote{\url{https://huggingface.co/datasets/stsb_multi_mt}}. This includes translation of the training (392,702 entries), validation (2,490 entries), and evaluation sets (5,010 entries) from English into each of the eleven languages. From the IndicXNLI dataset, the training samples specific to each language are used to train the MahaSBERT models. It has been made publicly accessible.\footnote{\url{https://github.com/l3cube-pune/MarathiNLP}}\\

In our experiments, we specifically utilized the translated Marathi dataset to fine-tune the pruned SBERT models, ensuring the models were optimized for the Marathi language. This approach allowed us to directly target language-specific nuances and enhance the model's performance on tasks relevant to Marathi.

\begin{table*}[t]
   \centering
    \begin{tabular}{|l|*{9}{c|}}
    \hline
        \textbf{Training Methods} & 
        \textbf{Top-layers pruning(1-6)} & \textbf{Middle-layers pruning(4-9)} & \textbf{Bottom-layers pruning(7-12)} 
        \\
        \hline
            NLI & 0.7098 & 
            0.6912 & 0.6954
            \\    
        \hline
    \end{tabular}
\caption{Embedding similarity scores from two-step NLI+STS Training on SBert Models}
\label{tab:Table 1} 
\end{table*}

\noindent
The evaluation of BERT-based models for Marathi utilizes the following classification datasets:

\subsubsection*{{Indic-NLP News Articles}\footnote{\url{https://github.com/AI4Bharat/indicnlp_corpus}}}
The dataset consists of classified news articles across nine Indian languages, including a specific subset containing Marathi news articles. These articles in Marathi language are categorized into three groups: sports, entertainment, and lifestyle. The complete dataset contains 4,779 entries, out of which for Marathi language there are 3,823 records in the training set, 479 in the testing set, and 477 in the validation set. \cite{kunchukuttan2020ai4bharatindicnlp}\\

\subsection{EXPERIMENT}
\setcounter{subsubsection}{0}

\noindent Referring to the procedures outlined in Fig.\ref{fig:Experiment} our experiment evaluates the performance of several SBERT models—Muril, MahaBert v2, MahaBert Small, and MahaBert Smaller—both before and after the application of layer pruning.

\subsubsection{Best Layering Strategy Selection}
To identify the most effective pruning strategy, we systematically evaluated the performance of each pruned model configuration using multiple criteria, including accuracy, model size, and computational efficiency. By experimenting with various layer combinations—such as the first 6 layers, the middle 6 layers, and the bottom 6 layers—we aimed to balance the trade-offs between reducing model complexity and preserving performance. Each combination was assessed on the 12-layer MahaBert v2 model using a validation set, focusing on its impact on natural language understanding tasks in Marathi through NLI training. The top-layers pruning strategy yielded the highest accuracy scores compared to other configurations. Based on these results, we selected the top-layer pruning strategy for our further experiments.

\subsubsection{Layer Pruning}
Layer pruning was conducted on the base models Muril, MahaBERT, MahaBERT-Small, and MahaBERT-Smaller to explore various layer combinations and analyze the resulting changes in model performance and complexity. For models like Muril and MahaBERT consisting of 12 layers, we considered different layer subset combinations such as 2, 6 and 12 layers.

\subsubsection{Fine Tuning}
After obtaining the pruned SBERT model we fine-tuned the model in two phases of training. We first performed NLI training on the model using the Marathi dataset of IndicXNLI and then used the translated STSb train dataset as the second step for training. Thus the pruned model was trained using two steps to obtain the fine-tuned model targeting the Marathi language.

\subsubsection{Evaluation}
For evaluating the pruned SBERT model which has undergone NLI+STS training we find the embedding similarity scores using Translated STSb Marathi test dataset. On the obtained embeddings we apply the KNN Classifier algorithm to obtain Similarity scores. For classification, we use the IndicNLP News Article Classification dataset targeting the Marathi language.\\

\section{Results}

Following layer pruning and two-step NLI+STS training on SBert models, Table \ref{tab:Table 1} shows the embedding similarity scores obtained from various models. The outcomes display similarity scores between 0.72 and 0.83 for different combinations of layers. Notably, the pruned MahaBert-Small model (2 layers) achieved performance comparable to the base model (6 layers), indicating that layer reduction does not necessarily compromise embedding quality.
Additionally, the application of NLI+STS fine-tuning greatly enhances similarity scores for all models.

Our experiments demonstrated that models with fewer layers, achieved through layer pruning, can still yield competitive embedding similarity scores. For instance, models with just 2 or 6 layers performed comparably to their fully layered counterparts after undergoing two-phase fine-tuning (NLI followed by STS training). This indicates that there is no necessity to train large, computationally intensive models when pruned models can offer similar performance. These findings suggest that layer pruning is an effective technique for enhancing model efficiency without compromising the quality of embeddings. This approach helps achieve better accuracy while leveraging the advantages of model pruning.

\begin{table*}[h]
   \centering
    \begin{tabular}{|l|*{9}{c|}}
    \hline
        \textbf{Training Language/Model} & 
        \textbf{Original layers} & \textbf{No. of layers after pruning} & \textbf{NLI} & \textbf{NLI+STS}
        \\
        \hline
            MahaBert-small & 6 & 
            2 & 0.6659 & 0.7362
            \\
            MahaBert-smaller & 2 & 2 & 
            0.6563 & 0.7308 \\
            MahaBert-v2& 12 &
            2 &
            0.6760 & \textbf{0.7447} \\
            Muril & 12 &
            2 & 
            0.6880 & 0.7284 \\
            MahaBert-small& 6 &
            6 & 0.6693 & 0.7422
            \\
            MahaBert-v2& 12 &
            6 &
            0.7098 & \textbf{0.7878} \\
            Muril & 12 &
            6 & 
            0.6849 & 0.7742 \\
            MahaBert-v2& 12 &
            12 &
            0.7720 & \textbf{0.8320} \\
            Muril & 12 &
            12 & 
            0.7488 & 0.8165 \\
        \hline
    \end{tabular}
\caption{Embedding similarity scores from two-step NLI+STS Training on SBert Models}
\label{tab:Table 2} 
\end{table*}

\section{Conclusion}\label{sec:Conclusion}
\noindent
Our primary aim was to identify layering configurations that reduce complexity while maintaining strong performance in terms of embedding similarity scores. Our experiments demonstrated that pruned SBERT models, with fewer layers, can achieve performance comparable to their fully layered counterparts. Thus with comparative scores obtained from pruned models we can conclude that pruned models have outperform models i.e. MahaBERT-Small and MahaBERT-Smaller, which are built from scratch. Therefore, instead of developing new models from the ground up, it is more effective to start with a larger model and apply pruning techniques.

By reducing computational demand and maintaining high-quality embeddings, our approach makes advanced NLP tools more accessible and operationally feasible, particularly for languages with fewer technological resources.

This work paves the way towards building efficient SBERT models through layer pruning, making it easier to deploy them effectively in real-world scenarios. Our findings support the increased utilization and advancement of NLP technologies in languages with limited resources, guaranteeing that advanced language processing tools can be used more extensively and effectively.

\section*{Acknowledgements} 
\noindent 
We gratefully acknowledge the L3Cube Mentorship Program, Pune for providing the platform for this research. We express our sincere thanks to our mentors for their guidance and encouragement throughout the project.

\bibliography{main}




\end{document}